\documentclass[conference]{IEEEtran}
\IEEEoverridecommandlockouts
\usepackage{cite}
\usepackage{amsmath,amssymb,amsfonts}
\usepackage{algorithmic}
\usepackage{graphicx}
\usepackage{textcomp}
\usepackage{xcolor}
\usepackage{multirow}
\usepackage{url}

\def\BibTeX{{\rm B\kern-.05em{\sc i\kern-.025em b}\kern-.08em
    T\kern-.1667em\lower.7ex\hbox{E}\kern-.125emX}}
\begin{document}

\title{Automatic Detection of Natural Disaster Effect on Paddy Field from Satellite Images using Deep Learning Techniques
}

\author{
\IEEEauthorblockN{Tahmid Alavi Ishmam, Amin Ahsan Ali, Md Ahsraful Amin, A K M Mahbubur Rahman}
\IEEEauthorblockA{
{Center for Computational \& Data  Sciences (CCDS), Independent University, Bangladesh}\\
}
\IEEEauthorblockN{tahmidalavi1999@gmail.com, aminali@iub.edu.bd, aminmdashraful@iub.edu.bd, akmmrahman@iub.edu.bd}
}

\maketitle

\begin{abstract}
This paper aims to detect rice field damage from natural disasters in Bangladesh using high-resolution satellite imagery. The authors developed ground truth data for rice field damage from the field level.  At first, NDVI differences before and after the disaster are calculated to identify possible crop loss.  The areas equal to and above the 0.33 threshold are marked as crop loss areas as significant changes are observed. The authors  also verified crop loss areas by collecting data from local farmers. Later,  different bands of satellite data (Red, Green, Blue) and (False Color Infrared) are useful to detect crop loss area. We used the NDVI different images as ground truth to train the DeepLabV3plus model. With RGB, we got IoU 0.41 and with FCI, we got IoU 0.51. As FCI uses NIR, Red, Blue bands and NDVI is normalized difference between NIR and Red bands, so greater FCI's IoU score than RGB is expected. But RGB does not perform very badly here. So, where other bands are not available, RGB can use to understand crop loss areas to some extent. The ground truth  developed in this paper can be used for segmentation models with very high resolution RGB only images such as Bing, Google etc.  
\end{abstract}

\section{Introduction}
Rice serves as the staple food for 135 million people in Bangladesh. Natural disasters can have a significant impact on rice production, and it is crucial to accurately identify the affected areas to take preventive measures or provide governmental aid to affected individuals. However, collecting such data requires significant human and economic resources. The use of satellite images, such as Sentinel-2 images, can automatically detect crop loss and segment the affected areas. In this research, we utilized NDVI as a ground truth and DeepLab V3 plus for semantic segmentation using both RGB and FCI images. The contributions of this study include identifying possible paddy loss areas, validating ground-truth data, and performing a comparative analysis of the performance using RGB and FCI images. Fig 1 is showing overview of our work.

\begin{figure}[!htb]
\includegraphics[scale=0.55]{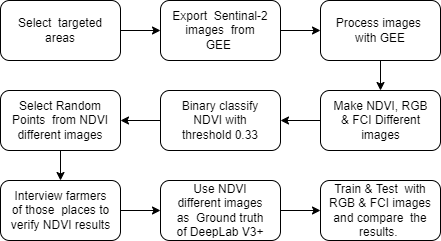}
\caption{Work flow}
\label{fig}
\end{figure}

\section{Background \& Context}
There have been works done on estimating rice production. They are estimated by NDVI value. Initially, we wanted to focus on boro paddy loss due to the heatwave, which occurred on 4th April 2021 [8]. But we cannot find any related work which shows estimating paddy loss due to heatwave from remote-sensing data. This mainly leads us to do this research. On the other hand, we annotated crop loss area using RGB image difference and FCI image difference as input in DeepLab V3 plus model. We also wanted to see how different band compositions performed at semantic segmentation to measure crop loss region.\par
From sentinal-2, combination images are used, which consists of multiple bands.  Combination imagery like False Color Infrared (FCI) is used for vegetation detection [3]. RGB satellite image was also combined to experiment and compare with FCI.  For Ground truth, we have used the Normalized difference vegetation index (NDVI).\par 

John Weier and David Herring (2000) first introduce NDVI as one of the measuring vegetation techniques [9].  NDVI is calculated by how much visible and near-infrared light is reflected from vegetation. The majority of visible light that strikes healthy plants is absorbed, whereas a considerable part of near-infrared light is reflected.  Vegetation that is unhealthy or sparse reflects more visible light but not as much near-infrared light.  Because Chlorophyll, a pigment found in plant leaves, absorbs visible light (between 0.4 and 0.7 m) and converts it to energy for photosynthesis.\par

On the other hand, the leaf cell structure reflects near-infrared light well (between 0.7 and 1.1 m).  These wavelengths of light are affected more by the number of leaves a plant has.  The NDVI value for a given pixel always ranges from minus one (-1) to plus one (+1); a zero value implies no vegetation, whereas a value near +1 (0.8-0.9) represents the highest density of green leaves conceivable [9].\par
In 2020's paper\,  the authors showed paddy rice phenology using Sentinel 2-A imaginary from NDVI band composition [5].  They verify greenness value with CCTV footage of the paddy field.  There they find that Sentinel Image 2-A can be used to estimate the paddy rice phenology, and the start and end of the paddy rice planting season can be determined using the NDVI greenness value.  They find out that at ndvi value  0.33, the first Phase of paddy vegetative started.  This NDVI value is very significant for us.  Because later, we will use this NDVI value as our threshold.\par

\section{Data set preparation}
\subsection{Study Area in Bangladesh}
According to the daily star's 9th April 2021 news, a massive nor'wester heat wave swept over the country on 4th April 2021 [8]. According to the Department of Agricultural Extension (DAE),  47,000 hectares of boro paddy(BRRI-29) have been affected in Kishoreganj, Netrakona, Mymensingh Sunamgnaj, Moulvibazar, Barishal and Patuakhali. We select our study area  in Sunamganj, Kisorganj \& Netrokona. These districts are side by side. According to the Department of Agricultural Extension 2020, Sunamganj has  219,300 hectares, Kishoreganj has  166,710 hectares and Netrakona has 184,530 hectares area where boro is cultivated, which is  73\%, 62\%, and 22\% respectively among haor districts [4].\par

\subsection{Source of satellite data}
Google Earth Engine (GEE) is a large publicly available geospatial dataset.We exported sentinal-2 images from GEE for our studied areas. Sentinal 2 carries an optical instrument payload which gives 13 spectral bands: four bands at 10 m, six bands at 20 m and three bands at 60 m spatial resolution [2]. We can combine these bands in various way for our task. 


As different channels capture different land textures, we use different combinations from Sentinel-2 data. These combinations are formulated in a certain way to accomplish specific tasks. In this research, RGB and FCI are the combinations of three bands. NDVI is a single band image that we calculate from other bands using a formula. Each index image has a specific characteristic that can detect specific classes. 


Because near-infrared (which vegetation strongly reflects) and red light (which vegetation absorbs), the vegetation index is good for quantifying the amount of vegetation. The formula for normalized difference vegetation index is (B8-B4)/(B8+B4); while high values suggest dense canopy [9]. \par



\subsection{Ground Truth making}
\subsubsection{Making NDVI Different images}
To make ground truth, we subtract 2nd image\'s NDVI band composition from 1st image\'s NDVI band composition. At Fig 2, we can see the the subtraction process. As we are conducting our study on three districts, each district has three years of data. So, we are getting nine NDVI different images. NDVI is a single band image for better visualization; we colored them using QGIS render type single-band pseudocolor. We colored NDVI values above 0.33 or equals as red and others as yellow which has shown at Fig 3. Later we exported our Rendered GeoTIFF images from Qgis.\par

\begin{figure}[htbp]
\centerline{\includegraphics[scale=0.5]{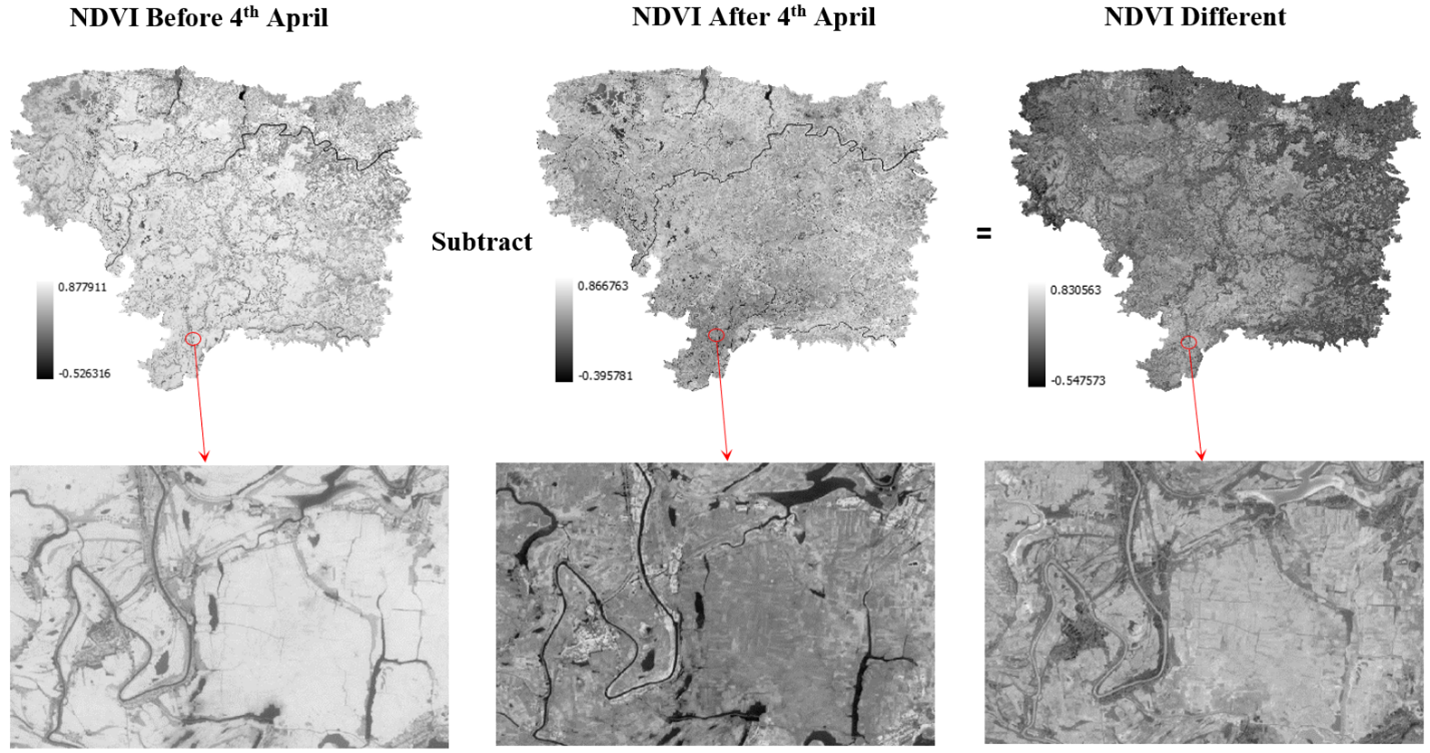}}
\caption{Illustration of NDVI Different. As an example, we took 2021's NDVI Before 4th April and NDVI After 4th April of Sunamganj. We can see that NDVI Before image is whitish than NDVI After which indicates there is some loss of vegetation at NDVI After}
\label{fig}
\end{figure}

\begin{figure}[htbp]
\centerline{\includegraphics[scale=0.5]{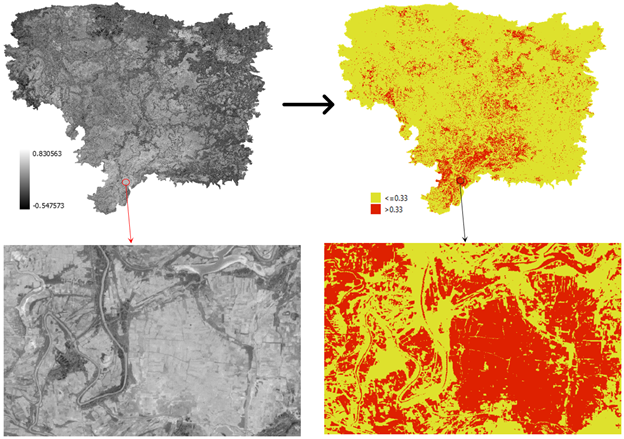}}
\caption{Coloring grayscale NDVI Different as the binary class where red means crop loss area \& yellow indicates the okay region}
\label{fig}
\end{figure}

At Fig 4 and Fig 5 we can see NDVI Different for Sunamganj \& Kisorgang.

\begin{figure}[!htb]
\centerline{\includegraphics[scale=0.5]{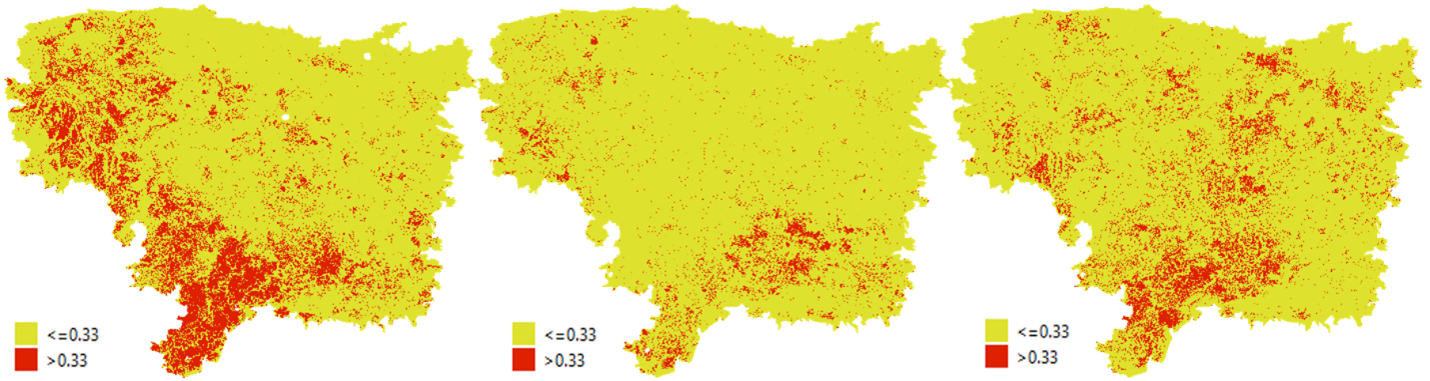}}
\caption{Sunamganj NDVI different from left to right 2019,2020 \& 2021}
\label{fig}
\end{figure}

\begin{figure}[!htb]
\centerline{\includegraphics[scale=0.5]{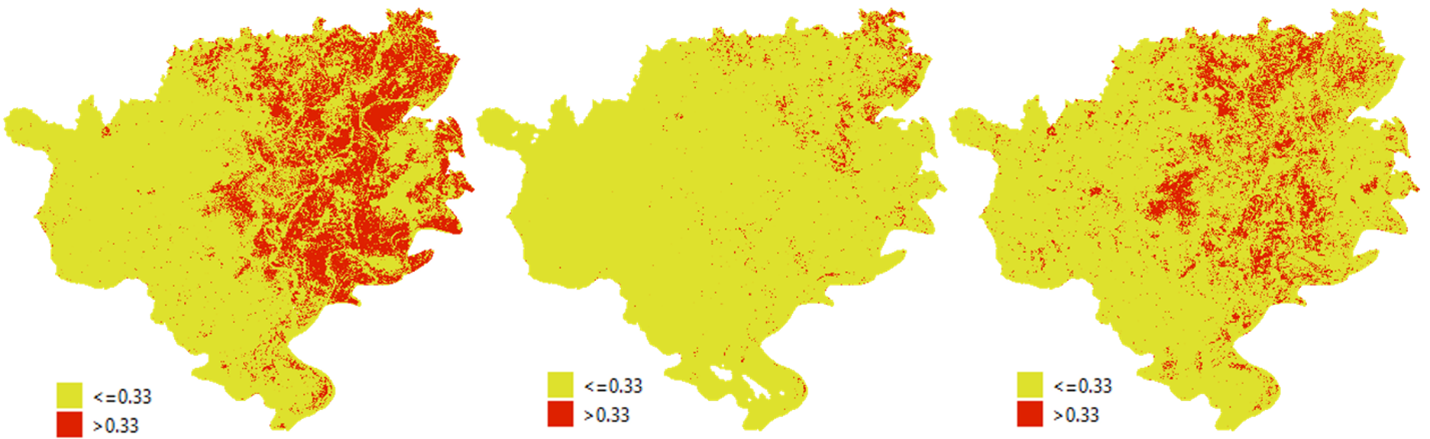}}
\caption{Kisorganj NDVI different from left to right 2019,2020 \& 2021}
\label{fig}
\end{figure}

As expected, we can see that in 2021 there is more NDVI difference than in 2020. However, here we can also see that in 2019 there is also a vast NDVI difference. It is because a 30-minute-long hailstorm struck south Sunamganj on 15-04-2019 morning [1]. Kisorhganj and Netrokona both are neighbour districts of Sunamganj; for that, we can see hailstorms impact on Kisorganj and Netrokona too. In the next subsection, we verify  these outcomes with field-level interview with corresponding farmers.\par

\subsubsection{Verify NDVI different as Ground Truth}
As we have seen from [5], NDVI can monitor rice phenology. But we need to be sure about our NDVI difference to use it as the ground truth of the deep lab v3+ model. For that, we need to go to each pinpoint location. \par
To do this, we selected two districts, Sunamganj and Kisorganj. Because we have observed similar NDVI patterns in Netrokona as well. On the other hand, Kishoreganj, Sunamganj, and Netrokona are located adjacent to each other. Suppose we can confirm Sunamganj and Kisorganj NDVI's different areas are rice field and affected due to hailstorms in 2019 and heatwave in 2021. In that case, we can also take Netrokona's NDVI different as correct and use all these NDVI different images as ground truth. \par
We randomly picked 30 points on Sunamganj and Kisorganj, as shown in Fig 6. Then by using their latitude and longitude, we found their address by reverse geocoding API. Here we use google map reverse geocode api \& barikoi reverse geocode api. \par

\begin{figure}[hbt!]
\centerline{\includegraphics[scale=0.35]{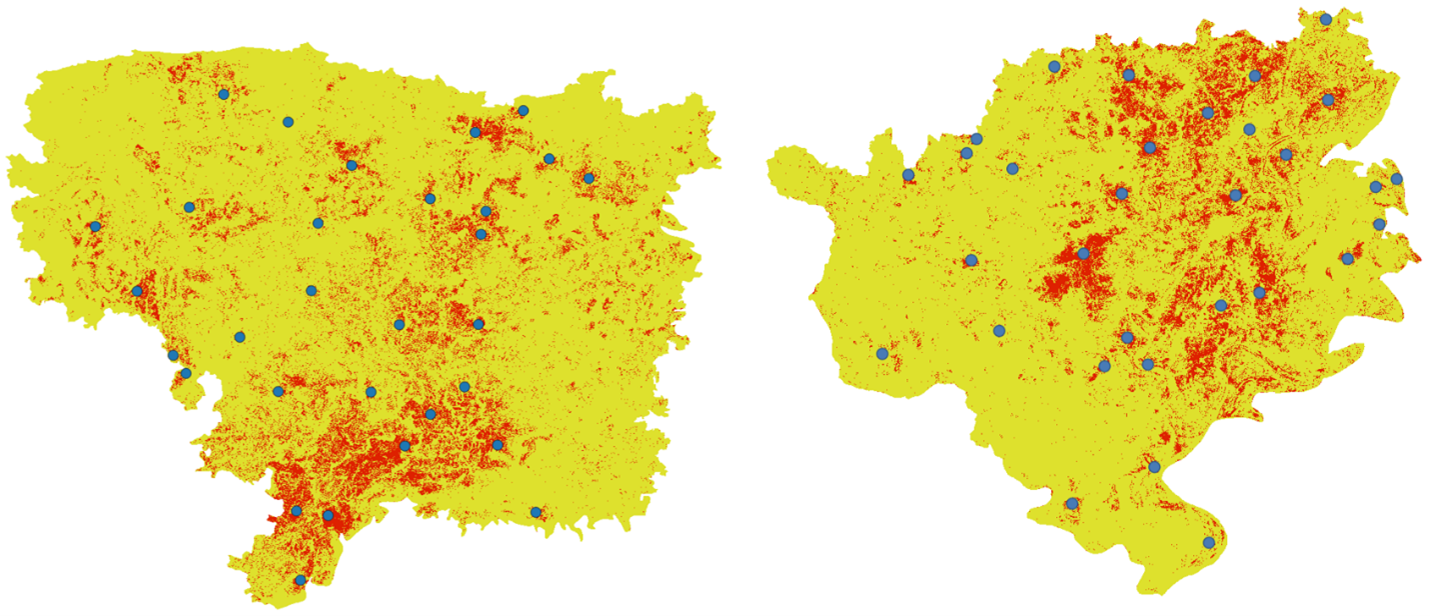}}
\caption{2021 before and after NDVI different of Sunamganj and Kisorganj with 30 sparsely selected points}
\label{fig}
\end{figure}

From these 30 places, we chose ten places to visit at Sunamganj and Kisorganj. 

We mainly focus on these questions at each point:
\begin{itemize}
\item Was there a paddy field? (This question was to verify if we can successfully recognize paddy field pattern from Sentinal-2A image)
\item If yes, then was that boro or another  variety of paddy?
\item Was the field of the circle affected by the heatwave in 2021 April and if yes, then how much?
\item Was the field of the circle affected by any natural disaster in 2020 and if yes, then how much?
\item Was the field of the circle affected by a hailstorm in 2019 and if yes, then how much?
\end{itemize}

At all places, we talked with farmers, and they gave us similar kind of information. We found that-
\begin{itemize}
\item All selected places were rice fields. They always cultivate rice there. At that time of the year, they grow Boro. 
\item For 2021 they all said they faced boro crop loss due to heatwave. They told their crop was burnt by hot air. 
\item In 2020 they did not face any vast kind of natural disaster and got very good Boro production. 
\item For 2019, most of them said about hailstorms. Even they told the hailstorm had a much more devastating effect on the crop than 2021's heatwave. 
\end{itemize}


From this ground level evaluation, we understand that our NDVI Different is indicating crop loss region correctly. \par

\section{Experimental setup}
\subsection{Export Sentinel 2 images from Google Earth Engine (GEE)}
To assess crop loss in Bangladesh after a heatwave on April 4, 2021, we need to collect sentinel-2 images before and after that date. However, clouds often cover Bangladesh's sky, and the sentinel-2 cloud mask is not effective for heavy cloud cover. Therefore, we must strategically set our date range. We collected all available images in GeoTIFF format from 2019, 2020, and 2021 for Sunamganj, Kisorganj, and Netrokona at a scale of 10 meters per pixel.

\subsection{Input image}
For the input image, we exported the RGB difference between before and after and the FCI difference between before and after from the GEE. The process of making  RGB different is demonstrated at Fig 7 and FCI different at Fig 8. After exporting the images from the GEE, we exported the rendered images with Qgis. We got a total of nine RGB difference images and nine FCI difference images. That means three RGB  difference images and three FCI  difference images for each district.\par

\textbf{RGB :} True color composite uses visible light bands red (B04), green (B03) and blue (B02) in the corresponding red, green and blue color channels, resulting in a natural-colored result [7].\par 
\textbf{FCI :} The false color infrared band combination
is meant to emphasize healthy and unhealthy vegetation. By
using the commbination of near-infrared (B8), red
(B04) and green (B03). Plants reflect near-infrared and green light while absorbing red. Plant-covered areas appear deep red due to their high near-infrared reflection, and denser plant growth is represented by darker red color.


\begin{figure}[!htb]
\includegraphics[scale=0.5]{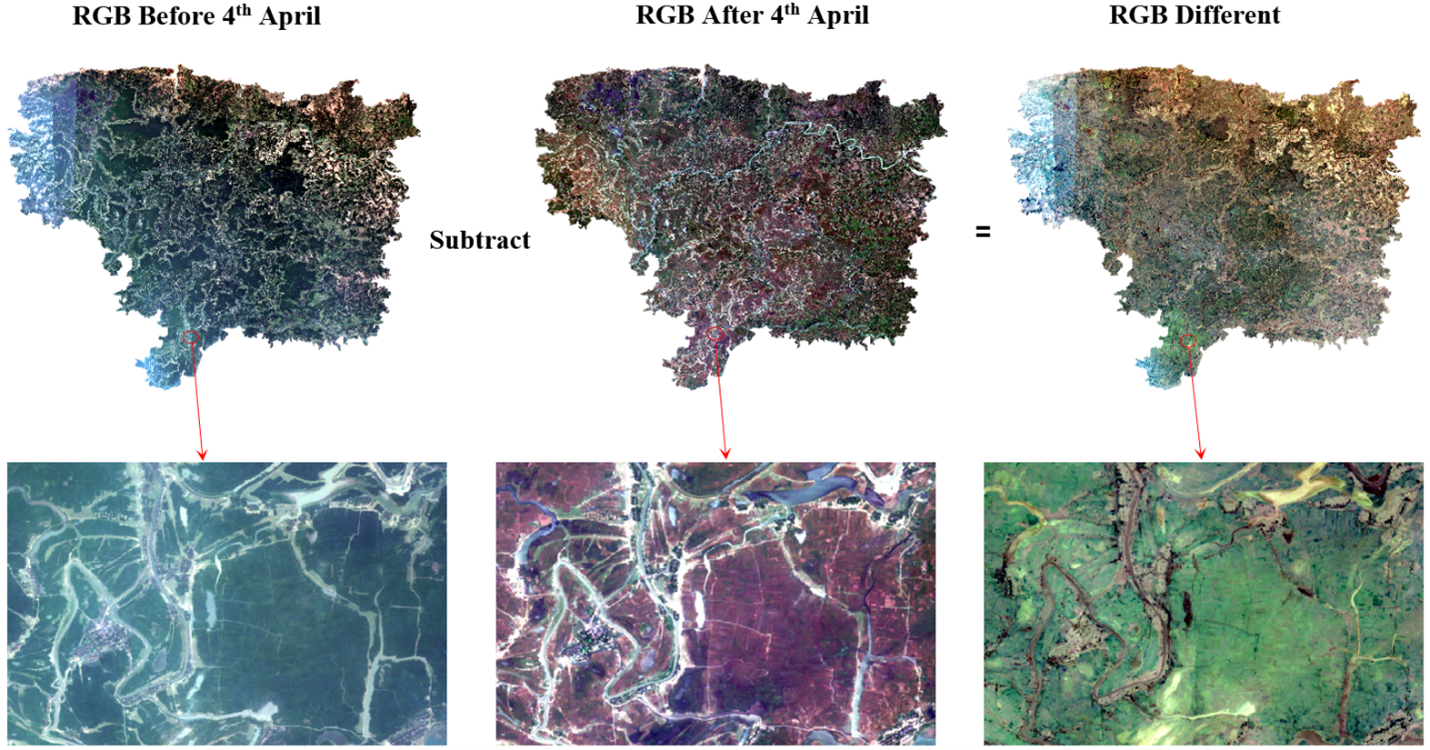}
\caption{Illustration of RGB Different. As an example, we took 2021's RGB Before 4th April and RGB After 4th April of Sunamganj. We can see that RGB Before image is greener than RGB After which indicates there is some loss of vegetation at RGB After}
\label{fig}
\end{figure}

\begin{figure}[!htb]
\includegraphics[scale=0.5]{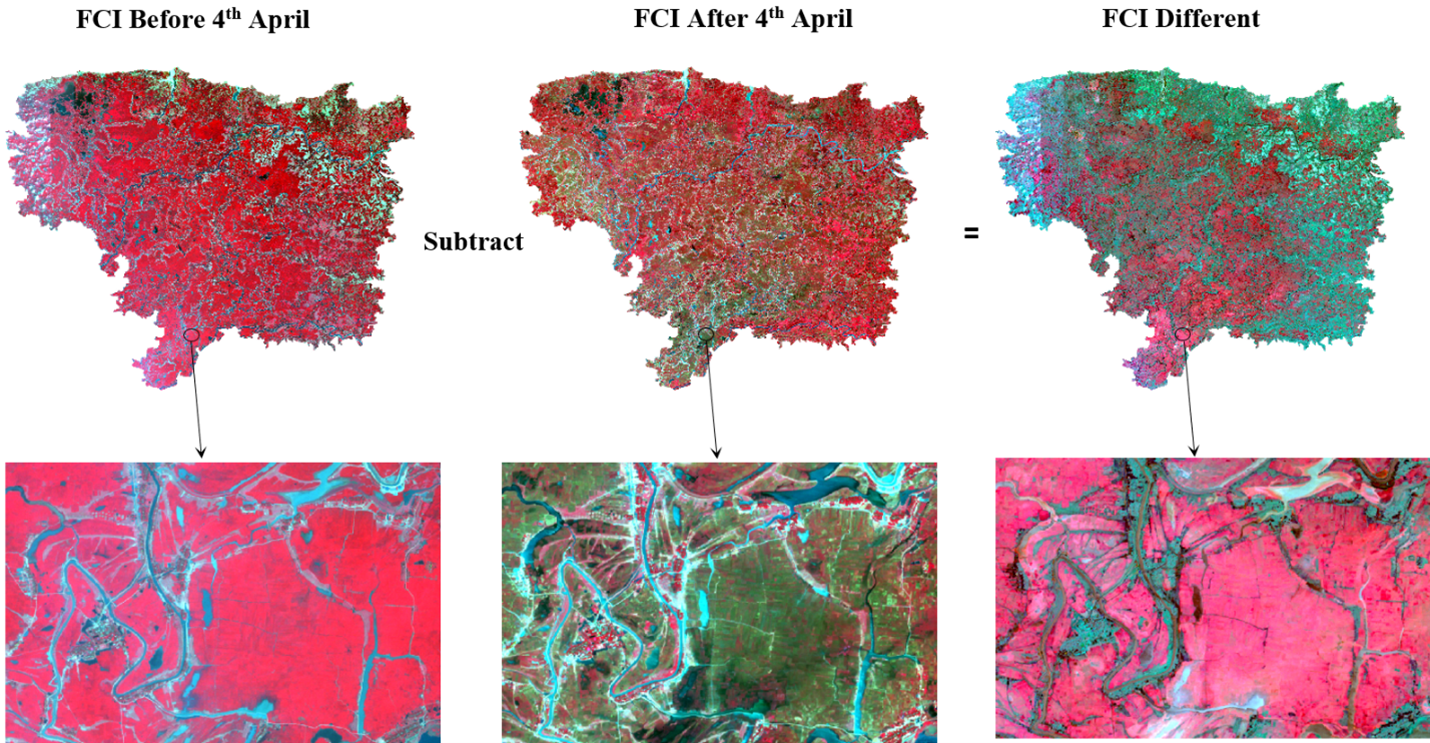}
\caption{Illustration of FCI Different. As an example, we took 2021's FCI Before 4th April and FCI After 4th April of Sunamganj. We can see that FCI Before image is redder than FCI After which indicates there is some loss of vegetation at FCI After}
\label{fig}
\end{figure}

\subsection{Pre-Processing}

\subsubsection{Split raster to feed model}
The original image is split into 256x256 to feed into the network.  As an example, Sunamganj's 2021 RGB different image's original dimension was 8987X7108. We zero-padded that for 256X256, which was 9216X7168. Then we split that image.\par
All split images were in GeoTIFF format. As we split our images, so from a single image, we have 1008 GeoTIFF for Sunamganj, 924 GeoTIFF for Netrokona and 810 GeoTIFF for Kisorganj. We have three years of data for each district. So in total, we got 3024 split images for Sunamganj, 2772 for Netrokona and 2430 for Kisorganj. We also mapped each split image with corresponding ground truth, in this case, which corresponds to split NDVI different GeoTIFF.\par

\subsection{Class Label}
We can see that our NDVI difference has mainly three colors. Black is in the background, red for the crop loss area and yellow for rest. So, for our segmentation, we have three classes. We declared our classes in the CSV file. In our model, we read these classes from that file. 
\subsection{Train, Validation and Test Splits}
We have three districts' data as each district has three years of data. So, we have a total of nine images. But we have split our images into 256X256. So now, we have a total of 8226 GeoTIFF images. We will use Sunamganj's data which is 36.76\% of our total dataset, as our train dataset, Netrokona's data which is 33.70\% of our total dataset, as our validation dataset \& Kisorganj data, which is 29.54\% of our total dataset as our test dataset.

\subsection{Evaluation Metrics}
\textbf{IoU :} We have calculated Intersect over Union (IoU) and from the model’s prediction and ground truths over test image pixels.  IoU is calculated by dividing the area of overlap by the area of union.\par

\begin{equation*}
IoU = \frac{Target \cap Prediction}{Target \cup Prediction} = \frac{TP}{TP + FP + FN}
\end{equation*}

\textbf{Mean IoU :} It means IoU average all over the classes. First, we calculated IoU by using the above formula for each class. In this case, for three classes. Then we divided by the number of classes. We used this at our testing result evaluation.\par

\textbf{Micro IoU :} Unlike mean Iou we calculate overall IoU for our model. We used micro IoU at training and validation phase. In this experiment, we calculated micro iou by using below formula.

\begin{scriptsize}
\begin{equation*}
\begin{split}
Micro IoU = \frac{Background TP + Loss TP + Ok TP + }{Background TP + Loss TP + Ok TP +} \\ {Background FP + Loss FP + Ok FP +} \\
{Background FN + Loss FN + Ok FN}
\end{split}
\end{equation*}
\end{scriptsize}

\subsection{Augmentation}
Though after splitting, we have lots of data for training. But from augmentation, we can achieve stronger generalization ability. We usually augmented our data by flipping, rotating, panning etc [6].. Here we use albumentations which is a fast and flexible image augmentation library.

\subsection{Post-Processing}
After getting output from Deeplab V3+, we set the same geotransform \& projection to output images as our input image. We do that by using python's Geospatial Data Abstraction Library (GDAL) and saving those split rasters. After all split image prediction, we merged all splitted images to one raster. 

\section{Experiment \& Result}
In our experiment, we used RGB and FCI difference images as input and applied augmentation to both the train input data and corresponding ground truth mask. Our loss function was dice Loss, and we used resnet101 as the encoder and softmax2d as the activation function. We used the Adam algorithm as the optimizer with a learning rate of 0.0001. The batch size was 16 for the training dataset and 1 for the validation dataset with 2 workers. We trained for 110 epochs, saving the model with better micro IoU score for the validation dataset. Training took approximately 4.30 hours on Colab pro. After training, we predicted and stored split test images, later merging them to obtain year-wise results.
We perform following experiments:
\begin{enumerate}
  \item Train, Validate, and  Test with RGB difference images.
  \item Train, Validate, Test with FCI difference images.
\end{enumerate}

In Fig 9 and Fig 10, we have presented the dice loss vs epochs graphs for both RGB and FCI images.\par
For RGB difference images, our best micro IoU score for the validation dataset was 0.9581 on epoch 87. At that time, the dice loss was 0.02386 and for the training dataset IoU score was 0.9432 and the dice loss was 0.02964. This was our saved model for RGB.

\begin{figure}[!htb]
\includegraphics[scale=0.52]{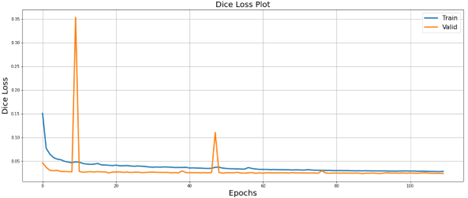}
\caption{Training vs Validation dice loss for RGB images}
\label{fig}
\end{figure}

At Fig 9, we can see some spikes for the validation dataset in RGB. As our batch size is 1  for validation, spikes are expected. We see these spikes particularly for RGB because RGB  images were noisier than FCI images.\par
For FCI difference images, our best IoU score for the validation dataset was 0.9693 on epoch 80. At that time loss was 0.01686 and for the training dataset IoU score was 0.9471 and loss was 0.02758. 


\begin{figure}[!htb]
\includegraphics[scale=0.52]{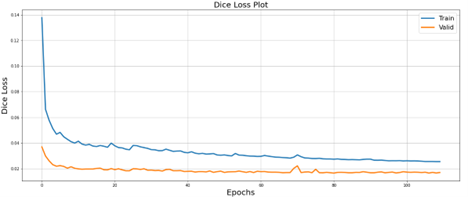}
\caption{Training vs Validation dice loss for FCI images}
\label{fig}
\end{figure}


The micro IoU score was a little better for FCI at the training phase. However, the difference is minimal.\par

At the testing phase, we have three years of data. First, we describe the overall performance for RGB and FCI difference images in test data; later, we analyze separate performance for each year’s image for test data. We provide the mean IoU and F1 here to compare the performance.\par

\begin{table}[!ht]
    \caption{Overall evaluation of RGB and FCI difference images in Test}
    \centering
    \begin{tabular}{|p{0.5cm}|p{0.7cm}|p{1.4cm}|p{1.0cm}|p{1.0cm}|p{1.0cm}|}
    \hline
    \textbf{Type} & \textbf{Matric} & \textbf{Background of the image} & \textbf{Crop compromised area} & \textbf{Rest of the areas} & \textbf{Mean}\\ \hline
            
    \multirow{2}{*}{RGB} & Mean IoU & 0.9991 & 0.4085 & 0.9045 & 0.7707  \\
    \cline{2-2}
    \cline{3-2}
    \cline{4-2}
    \cline{5-2}
    \cline{6-2}
    & F1 & 0.9995 & 0.5801 & 0.9498 & 0.8431 \\
    \hline \hline
    \multirow{2}{*}{FCI} & Mean IoU & 0.9994 & 0.5154 & 0.9162 & 0.8103  \\
    \cline{2-2}
    \cline{3-2}
    \cline{4-2}
    \cline{5-2}
    \cline{6-2}
    &  F1 & 0.9997 & 0.6802 & 0.9563 & 0.8787 \\
    \hline
\end{tabular}
\end{table}

\begin{figure}[!htb]
\includegraphics[scale=0.18]{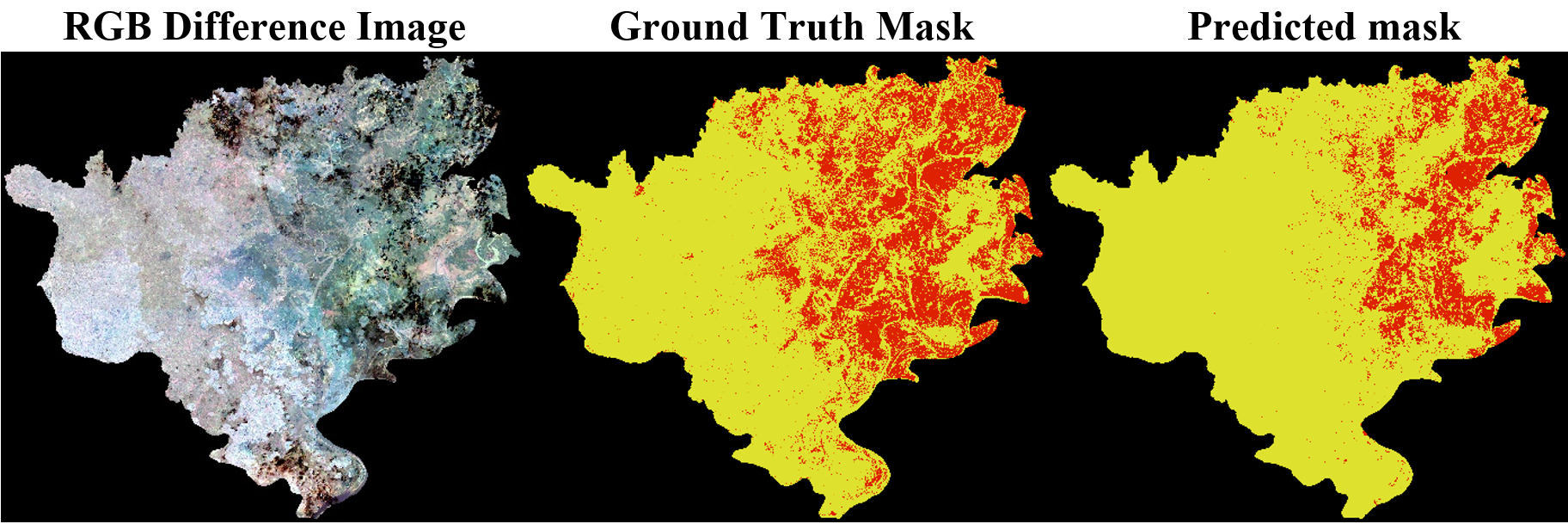}
\includegraphics[scale=0.18]{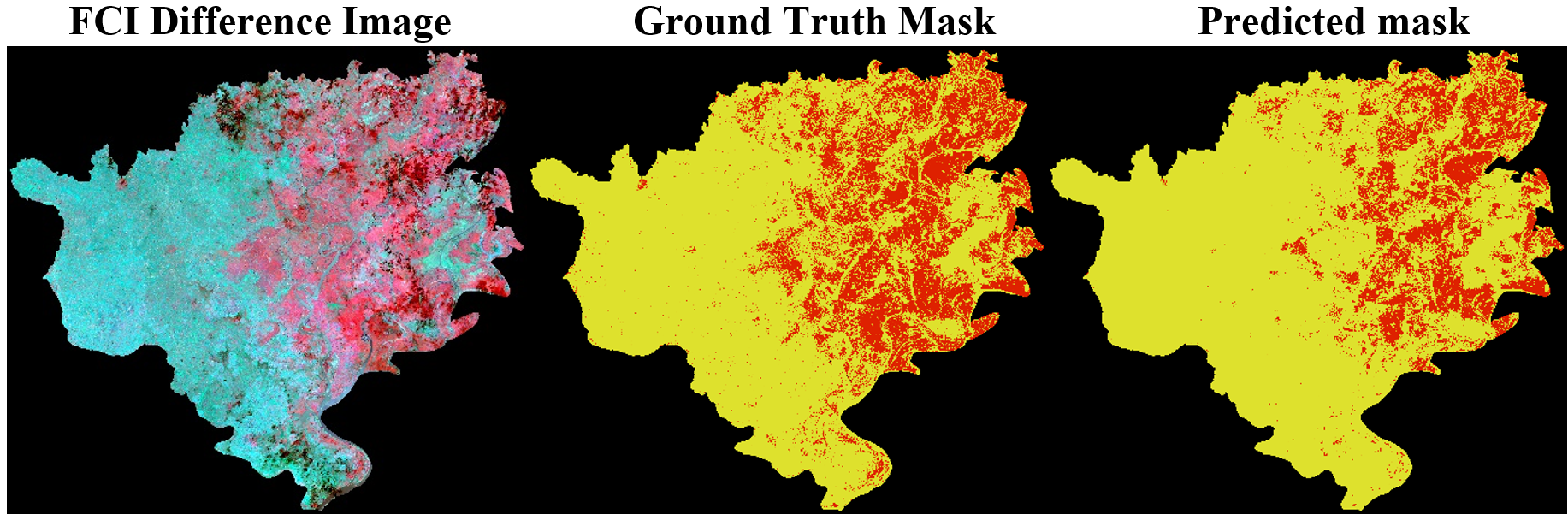}
\caption{2019's RGB and FCI difference images with ground truth mask and predicted mask}
\label{fig}
\end{figure}

\begin{table}[!ht]
    \caption{Evaluation of RGB and FCI difference images of Test (2019)}
    \centering
    \begin{tabular}{|p{0.5cm}|p{0.7cm}|p{1.4cm}|p{1.0cm}|p{1.0cm}|p{1.0cm}|}
    \hline
    \textbf{Type} & \textbf{Matric} & \textbf{Background of the image} & \textbf{Crop compromised area} & \textbf{Rest of the areas} & \textbf{Mean}\\ \hline
            
    \multirow{2}{*}{RGB} & Mean IoU & 0.9990 & 0.4936 & 0.8452 & 0.7792  \\
    \cline{2-2}
    \cline{3-2}
    \cline{4-2}
    \cline{5-2}
    \cline{6-2}
    & F1 & 0.9995 & 0.6609 & 0.9161 & 0.8588 \\
    \hline \hline
    \multirow{2}{*}{FCI} & Mean IoU & 0.9993 & 0.6140 & 0.8770 & 0.8301  \\
    \cline{2-2}
    \cline{3-2}
    \cline{4-2}
    \cline{5-2}
    \cline{6-2}
    &  F1 & 0.9996 & 0.7608 & 0.9344 & 0.8983 \\
    \hline
\end{tabular}
\end{table}

\begin{figure}[!htb]
\includegraphics[scale=0.18]{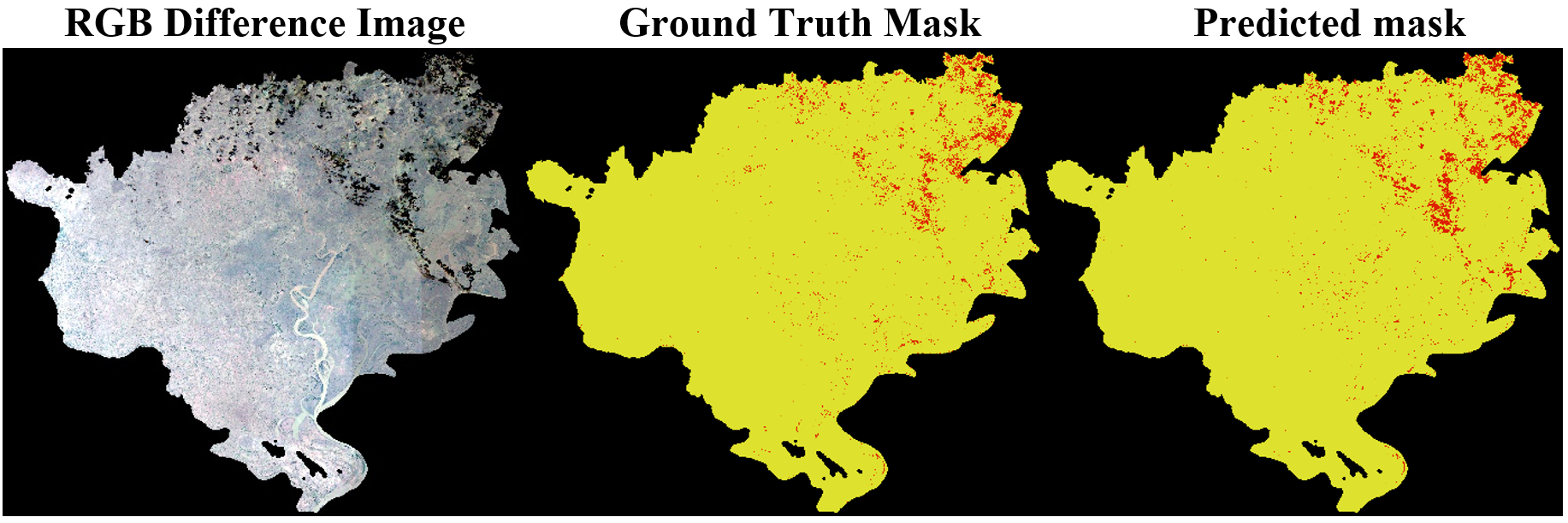}
\includegraphics[scale=0.18]{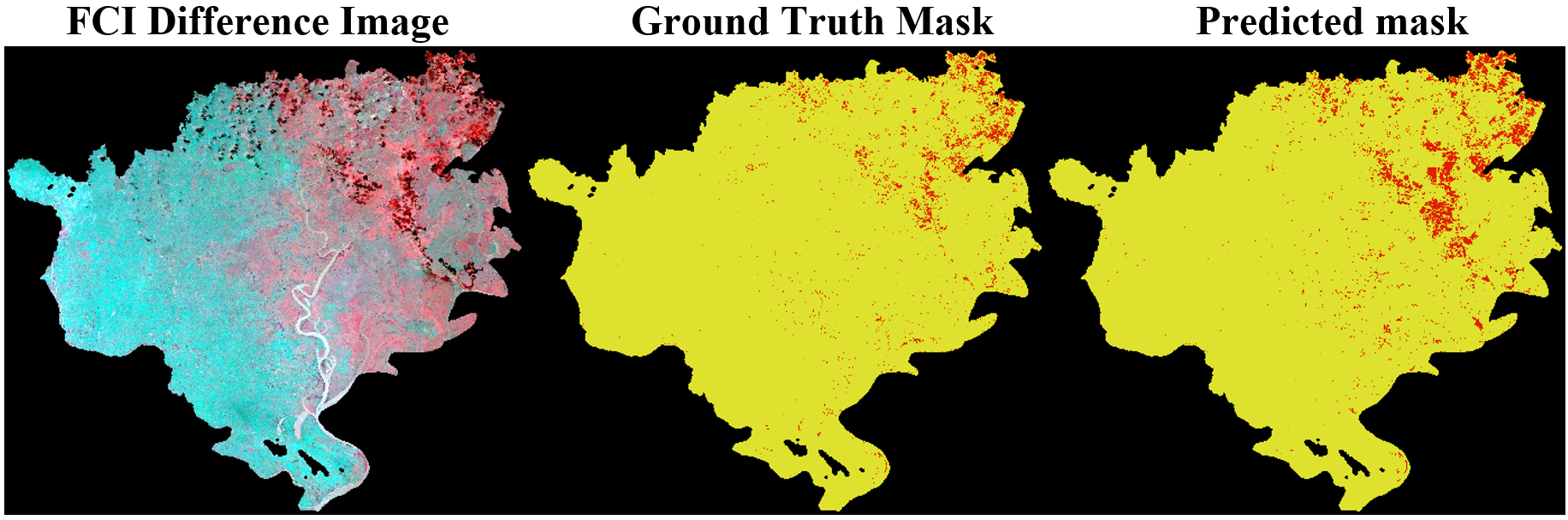}
\caption{2020's RGB and FCI difference images with ground truth mask and predicted mask}
\label{fig}
\end{figure}

\begin{table}[!ht]
    \caption{Evaluation of RGB and FCI difference images of Test (2020)}
    \centering
    \begin{tabular}{|p{0.5cm}|p{0.7cm}|p{1.4cm}|p{1.0cm}|p{1.0cm}|p{1.0cm}|}
    \hline
    \textbf{Type} & \textbf{Matric} & \textbf{Background of the image} & \textbf{Crop compromised area} & \textbf{Rest of the areas} & \textbf{Mean}\\ \hline
            
    \multirow{2}{*}{RGB} & Mean IoU & 0.9989 & 0.4459 & 0.9757 & 0.8068  \\
    \cline{2-2}
    \cline{3-2}
    \cline{4-2}
    \cline{5-2}
    \cline{6-2}
    & F1 & 0.9994 & 0.6168 & 0.9877 & 0.8680 \\
    \hline \hline
    \multirow{2}{*}{FCI} & Mean IoU & 0.9993 & 0.3998 & 0.9654 & 0.7881  \\
    \cline{2-2}
    \cline{3-2}
    \cline{4-2}
    \cline{5-2}
    \cline{6-2}
    &  F1 & 0.9996 & 0.5712 & 0.9824 & 0.8511 \\
    \hline
\end{tabular}
\end{table}

\begin{figure}[!htb]
\includegraphics[scale=0.18]{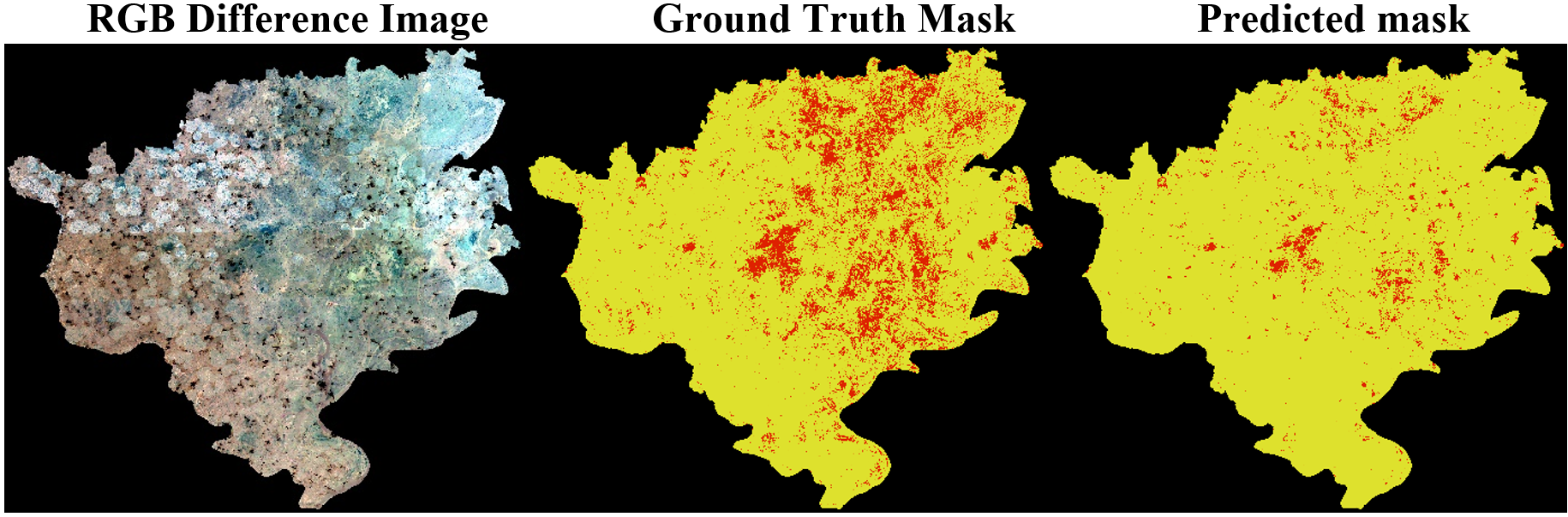}
\includegraphics[scale=0.18]{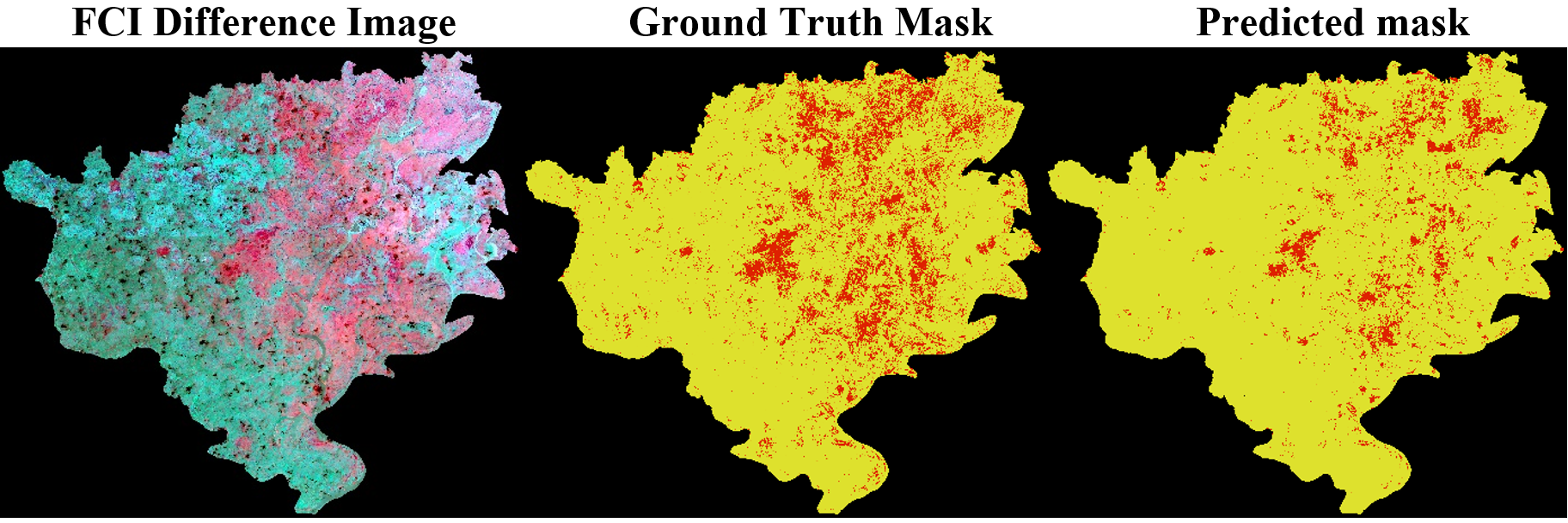}
\caption{2021's RGB and FCI difference images with ground truth mask and predicted mask}
\label{fig}
\end{figure}

\begin{table}[!ht]
    \caption{Evaluation of RGB and FCI difference images of Test (2021)}
    \centering
    \begin{tabular}{|p{0.5cm}|p{0.7cm}|p{1.4cm}|p{1.0cm}|p{1.0cm}|p{1.0cm}|}
    \hline
    \textbf{Type} & \textbf{Matric} & \textbf{Background of the image} & \textbf{Crop compromised area} & \textbf{Rest of the areas} & \textbf{Mean}\\ \hline
            
    \multirow{2}{*}{RGB} & Mean IoU & 0.9995 & 0.2414 & 0.8859 & 0.7090  \\
    \cline{2-2}
    \cline{3-2}
    \cline{4-2}
    \cline{5-2}
    \cline{6-2}
    & F1 & 0.9997 & 0.3890 & 0.9395 & 0.7761 \\
    \hline \hline
    \multirow{2}{*}{FCI} & Mean IoU & 0.9995 & 0.3850 & 0.9003 & 0.7616  \\
    \cline{2-2}
    \cline{3-2}
    \cline{4-2}
    \cline{5-2}
    \cline{6-2}
    &  F1 & 0.9997 & 0.5559 & 0.9475 & 0.8344 \\
    \hline
\end{tabular}
\end{table}


Table I shows that the mean Intersection over Union (IoU) score is 0.77 for RGB and 0.81 for FCI images. When specifically examining crop compromised areas, the IoU score is 0.41 for RGB and 0.52 for FCI. Analysis of data from Table IV reveals a very low IoU score for both FCI and RGB in compromised areas during the 2021 heatwave. Fig 13 shows low red pixels in the predicted masks for both FCI and RGB compared to the ground truth. However, Fig 12 indicates that there were many false-positive loss area pixels in the FCI during the 2020 season. Table III demonstrates that RGB performed better than FCI in crop compromised areas. During the hailstorm-affected year of 2019, the affected area's IoU score was greater than other years, with a score of 0.49 for RGB and 0.61 for FCI from Table II. In Fig 11, the predicted mask shows better detection of red areas in the FCI image compared to RGB. \par

\section{Conclusion}
This paper proposes an approach to develop ground truth data for paddy loss detection. 
It shows that performing Sentinal-2's NDVI subtraction before and after a disaster can be  a way to develop ground truth for paddy loss area. This method can be used to train various segmentation models for automatic segmentation of paddy loss area.  After the training,  RGB and FCI images seems effective for automatic segmentation of paddy loss areas. Though our IoU score was not very good. At the loss area's IoU, we got a better result for FCI than RGB. Another observation is, from the year-wise segmentation result, we found that result was not same for paddy loss due to heatwave in 2021 and paddy loss due to hailstorm in 2019. As we know, hailstorms and heatwaves affect paddy field in different ways, the different outcomes  are expected. However,  RGB \& FCI both do better for hailstorms. We can tell for very destructive disasters for paddy field like hailstorms, tornado, cyclone, flood; we can use  RGB in these cases. Our research is prone to heavy cloud-covered areas. In future, we will do further study to develop the segmentation model with Synthetic Aperture Radar (SAR) data that are not affected by clouds.\par

\section*{Acknowledgment}

This project is supported by Independent University Bangladesh and ICT Division of Bangladesh Government.

\end{document}